%% file: main.tex
\def\year{2022}\relax
\documentclass[letterpaper]{article} 
\input{preamble.tex}

\begin{document}
\maketitle

\input{abstract.tex}
\input{introduction.tex}
\input{background.tex}
\input{approach_methods.tex}

\input{empirical_methodology.tex}
\input{results.tex}
\input{conclusion.tex}

\section{Acknowledgements}
The research was partially funded by the Israeli Science Foundation grant \#2185/20

\bibliography{Bibliography}

\end{document}


\input{sections/appendix.tex}

%% file: preamble.tex
\usepackage{aaai22}  
\usepackage{times}  
\usepackage{helvet}  
\usepackage{courier}  
\usepackage[hyphens]{url}  
\usepackage{graphicx} 
\usepackage{natbib}  
\usepackage{caption} 
\DeclareCaptionStyle{ruled}{labelfont=normalfont,labelsep=colon,strut=off} 
\usepackage{algorithm}
\usepackage{algorithmic}

\usepackage{newfloat}
\usepackage{listings}
\floatstyle{ruled}
\newfloat{listing}{tb}{lst}{}
\floatname{listing}{Listing}
\title{``I Don't Think So'': Summarizing Policy Disagreements for Agent Comparison}
\author{
    Yotam Amitai, Ofra Amir\\ 
}
\affiliations{
    Faculty of Industrial Engineering, Technion I.I.T. \\
    
}

\usepackage{amsmath,amsthm,amssymb}
\usepackage{enumitem}
\usepackage[title]{appendix}
\setitemize{noitemsep,topsep=0pt,parsep=0pt,partopsep=0pt}
\setdescription{noitemsep,topsep=0pt,parsep=0pt,partopsep=0pt}

\newtheorem{definition}{Definition}

\newcommand{\disalg}[1]{DISAGREEMENTS}

%% file: abstract.tex
\begin{abstract}
With Artificial Intelligence on the rise, human interaction with autonomous agents becomes more
frequent. Effective human-agent collaboration requires users to understand the agent's
behavior, as failing to do so may cause reduced productivity, misuse or frustration. Agent strategy summarization methods are used to describe the strategy of an agent to its destined user through demonstration. A summary's objective is to maximize the user's understanding
of the agent's aptitude by showcasing its behaviour in a selected set of world states. While shown to be useful, we show that current methods are limited when tasked with comparing between agents, as each summary is independently generated for a specific agent. In this paper, we propose a novel method for generating dependent and contrastive summaries that emphasize the differences between agent policies by identifying states in which the agents disagree on the best course of action. 
We conduct user studies to assess the usefulness of disagreement-based summaries for identifying superior agents and conveying agent differences. 
Results show disagreement-based summaries lead to improved user performance compared
to summaries generated using HIGHLIGHTS, a strategy summarization algorithm which generates summaries for each agent independently.
\end{abstract}

%% file: introduction.tex
\section{Introduction}
With the maturing of reinforcement learning (RL) methods, RL-based agents are
being trained to perform complex tasks in various domains, including robotics,
healthcare and transportation. Importantly, these agents do not operate in a
vaccum  -- people interact with agents in a wide range of settings. Effective
interaction between an agent and a user requires from the latter the ability to
anticipate and understand its behavior. For example, clinicians should
understand treatment regimes recommended by agents to determine their viability.

To facilitate improved user understanding of agents' behavior, a
range of explainable RL methods have been developed
\cite{XRL_survey,heuillet2021explainability}. These can be divided into two explanation categories: (1) ``local'' explanation approaches explaining why an agent chose
a particular action in a given state, e.g., saliency maps highlighting the
information an agent attends to~\cite{greydanus2017visualizing}, and (2)
``global'' explanation methods that describe the policy of an agent more
generally, such as strategy summaries that demonstrate the agents' behavior in a
selected set of states~\cite{amir2019summarizing}. While these approaches
have been shown to improve people's understanding of agent behavior,
they are typically not optimized for a particular user task.

In this work, we aim to enhance users' ability to distinguish between the
behavior of different agents. Such scenarios arise when people are required to
select an agent from a set of alternatives. E.g., a user might need to choose
from a variety of smart home assistants or self-driving cars available on the
market. Importantly, there often isn't a clear ``ground-truth'' for which agent
is superior, as agents may prioritize alternative outcomes and users may occupy
different preferences. For example, some users may prefer self-driving cars that
value safety very highly, while others might be willing to relax safety
considerations to a degree to allow for faster driving. The ability to
distinguish policies is also important for model developers, as different
configurations of reward functions and algorithm parameters can lead to
different behaviors in unexpected ways, especially in domains where the reward
function is not obvious, such as healthcare \cite{gottesman2019guidelines}. 

One possible approach for helping users distinguish between policies of agents
is to use strategy summarization
methods~\cite{amir2018agent,amir2019summarizing}. 
Using these methods, a summary is generated for each agent, allowing the user to
compare their behavior. 
However, these approaches are not optimized for the task of agent comparison, as
each summary is generated \emph{independently}. For instance, the HIGHLIGHTS
algorithm~\cite{amir18highlights} selects states based on their importance as
determined by the differences in Q-values for alternative actions in a given
state. If two high-quality agents are compared, it is possible that they will
consider the same states as most important, and will choose the same (optimal)
action in those states, resulting in similar strategy summaries. In such a case,
even if the agents' policies differ in numerous regions of the state-space, it
will not be apparent from the summaries.

This work presents the \disalg~ algorithm which is optimized for comparing
agent policies.
Our algorithm compares agents by simulating them in parallel and noting
disagreements between them, i.e. situations where the agents' policies differ.
These disagreement states constitute behavioral differences between agents
and are used to generate visual summaries optimized towards showcasing the most
prominent conflicts, thus providing contrastive information. Our
approach assumes access to the agent’s strategy, described using a Markov
Decision Process (MDP) policy, and quantifies the importance of disagreements by
making use of agents’ Q-values.

To evaluate \disalg~, we conducted two human-subject experiment to test the following properties of the algorithm: Firstly, whether it improves users' ability to
identify a superior agent when one exists, i.e. given ground-truth performance measures. Secondly, do \disalg~ summaries outperform
HIGHLIGHTS summaries in conveying differences in behavior of agents to users.
Both
experiments make use of HIGHLIGHTS summaries as a baseline for comparison.
Results indicate a significant improvement in user performance for the agent
superiority identification task, while not falling short in the behaviour conveying task, as compared to HIGHLIGHTS.


Our contributions are threefold: \textit{i)} We introduce and formalize the
problem of comparing agent policies;  \textit{ii)} we develop \disalg~, an
algorithm for generating visual contrastive summaries of agents' behavioral
conflicts, and \textit{iii)} we conduct human-subject experiments, demonstrating
that summaries generated by \disalg~ lead to improved user performance compared
to HIGHLIGHTS summaries.

%% file: background.tex
\section{Related Work}
In recent years, explainable AI has regained interest, initially focusing mainly
on explaining supervised models. More recently, research has begun
exploring explanations of reinforcement learning agents~\cite{XRL_survey,
heuillet2021explainability}. 
In this work, we focus on global explanations that aim to describe the policy of
the agent rather than explain a particular action. Specifically, we develop a
new method for strategy summarization. In this section we describe in
more depth strategy summary methods~\cite{amir2019summarizing}.



Strategy summarization techniques convey agent behavior by demonstrating the
actions taken by the agent in a selected set of world states. The key question
in this approach is then how to recognize meaningful agent situations.

One such approach called HIGHLIGHTS~\cite{amir18highlights}, extracts
\emph{important} states from execution traces of the agent. Intuitively, a state
is considered important if the decision made in that state has a substantial
impact on the agent's utility. To illustrate, a car reaching an intended highway
exit would be an important state, as choosing a different action (continuing on
the highway) will cause a significant detour. 
HIGHLIGHTS has been shown to support people's ability to understand the
capabilities of agents and develop mental models of their behavior
~\cite{amir18highlights,Tobias}.

\citet{Sequeira2020} extended HIGHLIGHTS by suggesting additional importance
criteria for the summary state selection referred to as \emph{Interestingness
Elements}. \citet{huang2017enabling,lage2019exploring} proposed a different
approach for generating summaries based on machine teaching methods. The key idea underlying this approach is to select a set of states that is optimized to allow
the reconstruction of the agents' policy using imitation learning or inverse
reinforcement learning methods.


Common to all previous policy summarization approaches is that each summary is
generated specifically for a single agent policy, independent of other agents.
This can hinder users' ability to compare agents, as the
summaries might show regions of the state-space where the agents act similarly,
failing to capture useful information with respect to where the agent policies
diverge. For example, HIGHLIGHTS focuses on ``important'' states and it is
likely that the states found to be most the important to one agent will be
considered important by another agent as well. These could be inherently
important stages of the domain such as reaching the goal or evading a dangerous
state. If the agents act similarly in these important states, the HIGHLIGHTS
summaries of the agents might portray similar behavior, even for agents whose
global aptitude varies greatly. In contrast, if the summaries do differ from
one another and portray different regions of the state-space, they do not convey how
the alternative agent would have acted had it been tasked with the same
scenario.

To address these limitations, we propose a new approach that is specifically
optimized for supporting users' ability to distinguish between policies.  



\section{Background}
For the purpose of this work, we assume a Markov Decision Process (MDP) setting.
Formally, an MDP is a tuple $\langle S,A,R,Tr \rangle$, where S is the set of
world states; {A} is the set of possible actions available to the agent; $R:S
\times A \rightarrow \mathbb{R}$ is a reward function mapping each state, and
$Tr(s,a,s') \rightarrow [0,1] \; s.t. \; s,s' \in S, \; a\in A$ is the
transition function.
A solution to an MDP is a \emph{policy} denoted $\pi$. 

\paragraph{Summaries}



A summary, denoted $\mathbb{S}$, is a set of trajectories $T = \langle t_1,\dots t_n \rangle$.
Each trajectory $t$ being a sequence of $l$ consecutive states $t =
\langle s_i, \dots, s_D, \dots, s_{i+l} \rangle$
surrounding the disagreement state $s_D$ and extracted from the agent's
execution traces. 

We formally define a summary extraction process of an agent's policy given an
arbitrary importance function $Im$, mapping state-action pairs to numerical
scores.

\begin{definition}[Summary Extraction]
Given an agent's execution traces, a summary trajectory budget $k$, and an
importance function $Im$. \\
The agent's summary $\mathbb{S}$ is then the set of trajectories $T = \{
t_1,...,t_k \}$ that maximizes the importance function.
\begin{align}
    \mathbb{S} = \max_T \; Im(T)
\end{align}
\end{definition}
In this paper, our baseline is the HIGHLIGHTS algorithm, which computes
importance as a function of the $Q$-values in a given state. Specifically, we
implement the importance function from \citet{Tobias}, an extension to
HIGHLIGHTS, which suggests determining the importance of a state based on the
difference between the maximal and second $Q$-values.
Formally:
\begin{align}
   Im(s) = \max_a\; Q^\pi(a,s) - second\underset{a}highest\; Q^\pi(a,s)
\end{align}
The trajectory is then the sequence of states preceding and succeeding the
important state. 

%% file: approach_methods.tex
\section{Disagreement-Based Summaries}
We propose a new summary method which supports the comparison of alternative
agents by explicitly highlighting the \emph{disagreements} between them. Thus,
constructing \emph{contrastive} summaries that convey policy divergence between
agents exposed to the same world states. This approach is in line with the
literature on explanations from the social sciences, which shows that people
prefer contrastive explanations \cite{miller2018explanation}. We note that while
typically contrastive explanations refer to ``why not?'' questions and consider
counterfactuals, in our case the contrast is between the decisions made by two
different policies.

We next describe our proposed disagreement-based summary method. Specifically,
we formalize the notion of agent disagreement,
describe the ``\disalg~'' algorithm for generating a joint summary of the
behavior of two agents, and describe how to measure the importance of a disagreement state and disagreement trajectory.

\paragraph{Agent Disagreement}
The two main dimensions on which agents can differ are their valuations of
states and their policies, i.e. their choice of action. These dimensions are of
course related, as different state valuations will naturally lead to different
policies. Our definition of a disagreement focuses on the policy. We then utilize the value function for ranking purposes.

In other words, any state $s$ for which different agents choose different
actions is considered a disagreement state. We use these states to
analyze and portray how the agents differ from one another in their behavior.
Formally:
\begin{definition}[Disagreement State]
    Given two agents ${Ag}_{1}$ and ${Ag}_{2}$ with policies  $\pi_1, \pi_2$,
    respectively. Define a state $s$ as a disagreement state $s_D$ iff:
    \begin{align}
        \pi_1(s) \neq \pi_2(s)
    \end{align}
    The set of all disagreement states $\mathbb{D}$ would then be:
    \begin{align}
        \forall s \in S \; | \; \pi_1(s) \neq \pi_2(s) : s \in \mathbb{D}
    \end{align}

\end{definition}

For a compact MDP where every state may be computed, this definition would
suffice. Alas, for more complex settings containing a continuous or vast state
space, it is not feasible to compare all states. The proposed method must be
able to overcome this difficulty.

\paragraph{Identifying Agent Disagreements Through Parallel Online Execution}
Given two alternative agents to compare, we initiate an online execution of both
agents simultaneously such that we follow the first (denoted as the
\emph{Leader} or $L$ for short, with policy $\pi_{L}$), while querying the
second (denoted as the \emph{Disagreer} or $D$ for short, with policy $\pi_{D}$)
for the action \emph{it} would have chosen in each state. Both agents are
constrained to act greedily and deterministically with respect to their
$Q$-values. Upon reaching a disagreement state, we allow the \emph{Disagreer} to
``split-off'' from following the \emph{Leader} and continue independently for a
limited number of steps while recording the states it reaches for later
analysis. Once the limit is reached, we store the \emph{disagreement
trajectory}, and revert the \emph{Disagreer} back to the disagreement state,
from which it continues to follow the \emph{Leader} until the next disagreement
state is reached and the process is repeated.

\paragraph{The \disalg~ Algorithm}
The algorithm pseudo-code is supplied in Algorithm \ref{alg:disagreements}. It's parameters are summarized in Table \ref{tb:parameters}. 

\begin{table}[ht]
    \centering
    \small
    \resizebox{0.85\columnwidth}{!}{%
    \begin{tabular}{|p{1.6cm}|p{4cm}|p{0.5cm}|p{0.5cm}|}
    \hline
    \textbf{Parameter} & \textbf{Description} & \textbf{F} & \textbf{H}
    \\
    \hline
    $k$ & Summary budget, i.e. number of trajectories & 5 & 5 \\
    \hline
    $l$ & Length of each trajectory & 10 & 20 \\
    \hline
    $h$  & Number of states following $s$ to include in the trajectory &5& 10  \\
    \hline
    $numSim$ & The number of simulations (episodes) run by the \disalg~ algorithm &10&10  \\
    \hline
    $overlapLim$ & Maximal number of shared states allowed between two trajectories in the summary & 3& 5 \\
    \hline
    $impMeth$  & Importance method used for evaluating disagreements & \emph{Last State} & \emph{Last State}\\
    \hline
    \end{tabular}}
    \caption{Parameters  for Frogger \& Highway domains.}
    \label{tb:parameters}
\end{table}

\emph{The Algorithm.} First, three lists are initialized for the Leader traces,
disagreement states and Disagreer trajectories (lines 4--6). Then, simulation of the agents are run (lines 7--27). Each simulation collects all
states seen by the Leader during the execution (line 24), disagreement states
(line 13), and the Disagreer trajectories (lines 14--19). Each step of the
simulation, both agents are queried for their choice of action (lines
10--11), if they disagree on the action --- a disagreement state is added
(line 13) and a disagreement trajectory is created (lines 14--19), after which
the simulation is reverted to the last disagreement state (line 21). Upon simulations completion, all disagreement trajectories (coupled pairs of
Leader and Disagreer trajectories) are obtained (line 28) and the most important ones are passed as output (line 29).

\begin{algorithm}[tb]
    \caption{The \disalg~ algorithm. }
    \label{alg:disagreements}
\begin{algorithmic}[1]
    \STATE {\bfseries Input:} $\pi_L,\pi_D, k, l, h,$ \STATE $overlapLim,
    numSim, impMeth$ \STATE {\bfseries Output:} $\mathbb{S}$ \STATE $L_{Tr}
    \leftarrow$ empty list \textit{\;\;\;\#Leader traces} \STATE $\mathbb{D}
    \leftarrow$ empty list \textit{\;\;\;\#Disagreement states} \STATE $D_T
    \leftarrow$ empty list \textit{\;\;\;\#Disagreer trajectories}  \FOR
    {$i=1$ {\bfseries to} $numSim$} \STATE $sim, s = InitializeSimulation()$
    \WHILE {$(!sim^{\pi_L}.ended())$} \STATE $a^{\pi_L} \leftarrow
    sim.getAction(\pi_L(s))$ \STATE $a^{\pi_D} \leftarrow
    sim.getAction(\pi_D(s))$ \IF{$a^{\pi_L} != a^{\pi_D}$} \STATE
    $\mathbb{D}.add(s)$ \STATE $d_t \leftarrow$ empty list
    \textit{\;\;\;\#Disagreer trajectory} \FOR{$i=1$ {\bfseries to} $h$}
    \STATE $s^{\pi_D} \leftarrow sim.advanceState(\pi_D)$ \STATE $a^{\pi_D}
    \leftarrow sim.getAction(\pi_D(s))$ \STATE $d_t.add(s)$ \ENDFOR \STATE
    $D_T.add(d_t)$ \STATE $sim, s = reloadSimulation(D_s[-1])$ \ENDIF \STATE $s
    \leftarrow sim.advanceState(\pi_L)$ \STATE $L_{traces}.add(s)$ \ENDWHILE
    \STATE $runs = runs+1$ \ENDFOR \STATE $DA_T \leftarrow
    disagreementTrajPairs(\mathbb{D},L_{Tr}, D_T, l, h)$ \STATE $\mathbb{S}
    \leftarrow topImpTraj(DA_T, k, overlapLim, impMeth)$
    \end{algorithmic}
\end{algorithm}

Since the \emph{Leader} agent controls which areas of the domain state-space are reached, we repeat the process again, reversing the agent's roles. This is important because the Leader determines which regions of the state-space are reached and as such also conveys information about the agent's preferences. This process results in two sets of disagreements states (one for each agent as the \emph{Leader}). Typically, these sets can be very large, and it is infeasible for a user to explore all of them. 
Therefore, a ranking procedure is
necessary for the disagreements found in order to generate a compact summary. We next describe approaches for
quantifying the importance of a disagreement.



\paragraph{Disagreement Importance} 
Various methods can be used for determining disagreement importance. We first define the notion of a state's value based on multiple
agents.

\textbf{State Value}
We assume the agents are $Q$-based, possessing a $Q$ function ($Q(s,a)
\rightarrow \mathbb{R}$) which quantifies
their valuation of state-action pairs, denoted as $Q$-values. $Q$-values are calculated and adjusted during the
training phase of the agent and depend on the algorithm used as well as on the
specification of the reward function. Therefore, $Q$-values of different agents
may vary greatly. This renders each individual agent's assessment of a
state-action pair as its own estimate rather than representing a ground truth.
The values themselves may not even be on the same scale. To allow for comparison
between values, we normalize each agent's $Q$-values by dividing by the maximum
$Q$-value appearing in each, thus rendering each value an indication of how good
a state-action pair is compared to the best one observed by the agent. Formally:
\begin{align}
    Q' = \frac{Q-\min_{s,a} \big(Q(s,a) \big)}
    {\max_{s,a} \big (Q(s,a) \big )- \min_{s,a} \big(Q(s,a) \big)}  
\end{align}

Since our agents' action selection is greedy and deterministic with respect to
their $Q$-values, we denote the value of state $s$ as the highest
$Q$-value associated with it.
 \begin{definition}[Agent State Value]
    Given an agent $Ag$, its $Q$ function $Q^{Ag}$ and a state $s$, we define
    the value of $s$ as:
    \begin{align}
       V^{Ag}(s) = \max_a {Q'}^{Ag}(s,a)
    \end{align}
 \end{definition}

Alas, measuring only the importance of a single state has its limitations.
Without ground truth, we are left only with the agents' estimations which may be
flawed. Suppose our agents reach a disagreement state where both are convinced
the other's action is a poor choice. They each go their separate ways only to
reunite at a shared state after several steps with minimal to no impact on the
succeeding execution, e.g. overtaking a vehicle from the left or from the
right. This realization led us to formulate a trajectory-based approach for
determining the importance of a disagreement state. 

\paragraph{Disagreement Trajectory Importance}
To determine the importance of a disagreement state, we compare the trajectories
that branch out of it by following each agent. We formulate trajectory value
metrics to evaluate these, while constraining ourselves to observing only
trajectories of similar length.

A trajectory ${t}_{h}^{\pi}(s) = \langle s_{+1}, ..., s_{+h} \rangle$ denotes
the sequence of states encountered when following state $s$ for $h$ steps
according to a policy $\pi$. Since each agent evaluates states differently, we
consider the value of a state as the sum of both agents' valuations, i.e. $V(s)
= V^{L}(s) + V^{D}(s)$. There are numerous ways to quantify the importance of a
trajectory. A description of several methods tested in our work is provided in
the Appendix. The summaries generated for the user studies made
use of the \emph{last-state} importance method.

\textbf{Last State Importance:}
We define the importance of a disagreement
trajectory as the difference between the values of the last states reached by
each of the agents. This reflects how ``far off'' from each other the
disagreement has led the agents. Formally:
\begin{align}
    Im({t}_{h}^{\pi_{L}}(s),{t}_{h}^{\pi_{D}}(s)) = 
    |V(s^{\pi_L}_{+h}) - V(s^{\pi_D}_{+h})|
\end{align}
This metric achieves a high importance score when one agent arrives at a state
which both agents agree is valuable, while the other agent arrives at a state whose value both agents agree is poor.

\paragraph{Trajectory Diversity}
Using these methods we are able to acquire the set of disagreement states
$\mathbb{D}$ ordered by importance, and for each, their corresponding
trajectories. These shall be woven together to create a visual summary for
displaying to the user. To increase the coverage of the summary and avoid
showing redundant trajectories (in both methods), we restrict the summary generated to not contain
\emph{i)} multiple trajectories that end or begin at the same state, \emph{ii)}
trajectories where the Leader and Disagreer share the same state before-last
and \emph{iii)} overlapping trajectories which share more than a predefined
number of states.

%% file: empirical_methodology.tex
\section{Empirical Methodology} 
\label{sec: emp}

To evaluate our method we conducted two user studies. The first was designed to
assess whether \disalg~ summaries help users identify the superior between two alternative agents, while the second user study examines whether such
summaries are useful for conveying agent behavior differences. In both studies we
use the HIGHLIGHTS algorithm as a baseline for comparison. We note here that
there is a significant difference between the output summaries of both methods
rooted in the fact that \disalg~ was designed for presenting two policies in a contrastive manner. This is achieved by portraying both agents simultaneously on the screen.
To our knowledge no other global explanation methods exist that directly compare
 policies and visualizes their differences to lay users.

\emph{Empirical Domains.} To evaluate our algorithm we generated summaries of
agents playing the game of Frogger \cite{Frogger_game} and controlling a vehicle
in a highway environment \cite{highway-env}. 

\textbf{Frogger} The objective of the game is to guide a frog from the bottom of
the screen to an empty lily-pad at the top of the screen. The agent controls the
frog and can initiate the following four movement actions: up, down, left or
right, causing the frog to hop in that direction. To reach the goal the agent
must lead the frog across a road with moving cars while avoiding being run over,
then, the agent must pass the river by jumping on passing logs. This domain
allows us to compare different agents in a setting with ground truth information
about agents' skill, i.e. game score.

\textbf{Highway} This domain consists of a busy highway with multiple lanes and
vehicles. The
agent controls a vehicle driving through traffic with the intent of avoiding
collisions. The agent can choose to move right or left (changing lanes),
increase or decrease velocity or stay idle, i.e. make no change. There is no
defined target the agent is required to reach, instead the road goes on
continuously. This property allows us to observe the agent's general behavior and preferences
instead of focusing on its progression towards reaching the goal.Screenshots from the \disalg~ output summaries displayed in Figure \ref{fig:
domains}.

\begin{figure}[ht]
	\centering
    \frame{\includegraphics[width=0.7\columnwidth]{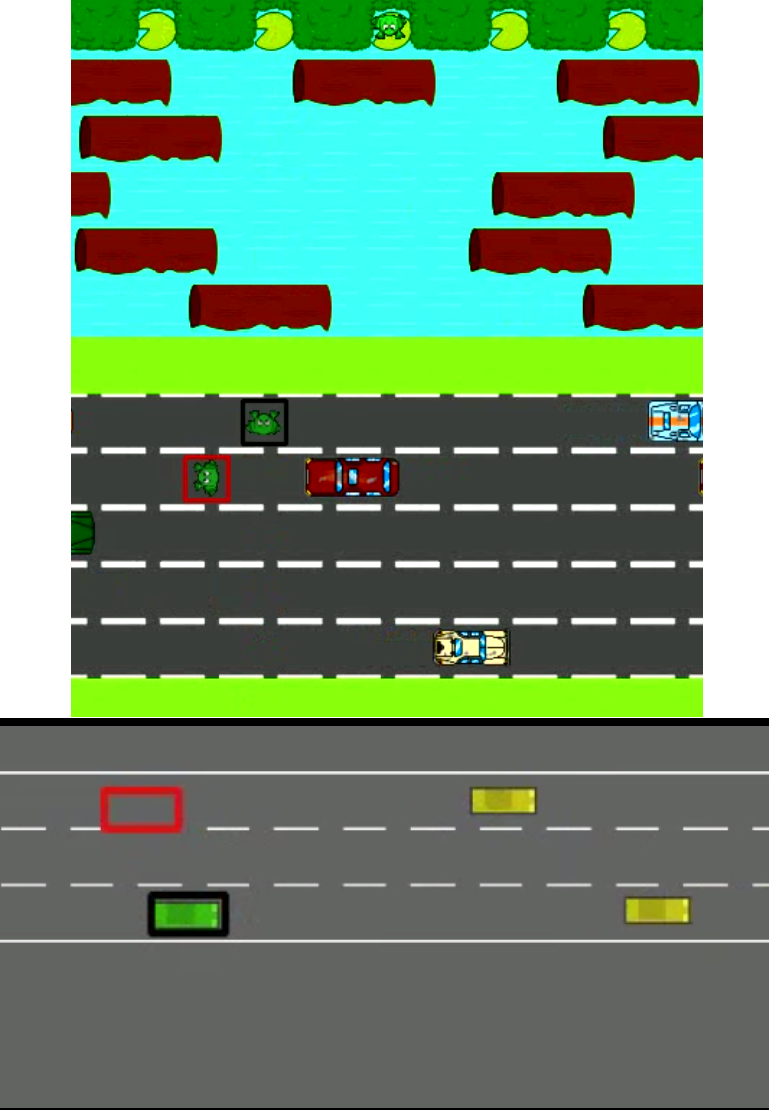}}\\
	\caption{\textbf{Top}: Frogger; \textbf{Bottom}: Highway.\\ Red \& black rectangles represent different agents.}
	\label{fig: domains}
	\vspace{-0.3cm}
\end{figure}

\emph{Frogger Agents.} We made use of the framework developed by
\citet{Sequeira2020} to test the \disalg~ algorithm on multiple configurable
agents of varying capabilities. Three different agents were trained using standard $Q$-learning \cite{watkins1992q}, based on the configurations provided by the framework. 
\begin{itemize}
	\item \textbf{Expert (E):} 2000 training episodes. Default rewards. Average
	game score: 110,000.
	\item \textbf{Mid-range (M):} 1000 training episodes. Default rewards.
	Average game score: 50,000.
	\item \textbf{LimitedVision (LV):} 2000 training episodes. Default rewards.
	Lower perception of incoming cars. Average game score: 55,000.
\end{itemize}
Agent performance was calculated by averaging the game score of ten
executions. Each agent's unique configuration contributes to its performance,
providing us a ground truth for the assessment tasks we present to the
experiment participants. The agents' skill hierarchy, based on their average
score, is as follows: $E>LV>M$. An important requirement for the
experiment was that all agents have a decent ability to play Frogger.
Prior to this study, we verified via an additional experiment that HIGHLIGHTS summaries are indeed useful for comparing Frogger agents that differ \emph{substantially} in their skills (see Appendix).
All HIGHLIGHTS and \disalg~ summaries were generated for fully trained agents, thus
reflecting their final policies.

\emph{Highway Agents.}  Agents with varying behaviors were trained by altering their reward functions. All highway agents were trained for 2000 episodes using double DQN architecture
\cite{hasselt2010double} and rewarded for avoiding collisions.
\begin{itemize}
	\item \textbf{ClearLane (CL):} Rewarded for high velocity while maximizing the distance between
	itself and the nearest vehicle in front of it.
	\item \textbf{SocialDistance (SD):} Rewarded for maximizing the distance
	between itself and the closest $k$ vehicles.
	\item \textbf{FastRight(FR):} Rewarded for high velocity and driving in the
	rightmost (bottom) lane.
\end{itemize}
Henceforth,
we will refer to all domain agents by their abbreviations.

\emph{Summary Attributes}
All summaries were composed of five trajectories made up of sequential states, ten for Frogger and twenty for Highway. These contained the important state at the center of the
trajectory, with half the states preceding and the rest succeeding it.
Video-clips of the summaries were generated to present to the users and a
fade-in and fade-out effect was added to further call attention to the
transition between trajectories.
 For more details, sensitivity analysis and the complete surveys, see Appendix.

\paragraph{Experiment 1 - Identifying Superiority}
The objectives of the first experiment were twofold. Firstly, to support our
claims regarding the limitations of the HIGHLIGHTS algorithm for comparing
agents, and secondly, to compare the \disalg~ algorithm to HIGHLIGHTS and show
its added value. 

\emph{Hypotheses.} 
We hypothesized that summaries generated by the HIGHLIGHTS
algorithm are limited in their ability to help users distinguish between agents
and the \disalg~ algorithm is more suited for this task. More specifically,
we state the following hypotheses: \\
 \textbf{H1:} Participants shown summaries generated by HIGHLIGHTS for
 agents of decent skill will struggle to identify the better performing
 agent. \\
\textbf{H2:} Participants shown summaries generated by the \disalg~
algorithm will exhibit a higher success rate for identifying the better
performing agent, compared to ones shown HIGHLIGHTS summaries. 


\emph{Experimental Conditions}
A between-subject experimental setup was designed with two experimental
conditions that varied in the summary generation method, \disalg~ or
HIGHLIGHTS~\cite{amir18highlights}. Participants were randomly assigned a
condition.

\emph{Participants.} 74 participants were recruited through Amazon Mechanical
Turk (27 female, mean age $= 36.51$, STD $= 9.86$), each receiving $\$3$ for
their completion of the Task. To incentivize
participants to make an effort, they were provided a bonus of 10 cents for each
correct answer in the superiority identification task.
Participants who spent less than a threshold duration of time on experiment tasks, based on the length of the task summary video, were filtered out.




\emph{Procedure.} Participants were first introduced to the game of Frogger and
the concept of AI agents. Each explanation was followed by a short quiz to
ensure understanding before advancing to the task. Next, participants were
randomly split into one of two conditions and were shown summary videos of pairs
of different agents generated using either \disalg~ or HIGHLIGHTS. 

Participants in both groups were first introduced to the summary method they
would be shown and were required to pass a quiz to ensure their understanding.
Participants were then asked to choose the better performing agent based on the
summary videos. They were able to pause, play and repeat the summary videos
without restrictions, allowing freedom to fully inspect the
summary before deciding which agent they believe is more skillful. Participants
were also asked to provide a textual explanation for their selection and to rate
their decision confidence on a 7-point Likert scale (0 - not at all confident to 6 - very confident). Overall, there were 3 pairs of
agent comparisons $\langle E,M\rangle \langle E,LV\rangle \langle LV,M \rangle
$. The ordering of the agent pairs was randomized to avoid learning effects, and
participants were also not told if the same agent appeared in multiple
comparisons, that is, they made each decision independently of other decisions.

Participants in the HIGHLIGHTS condition were shown a HIGHLIGHTS summary of each
agent (i.e. two separate videos, one for each agent.), while participants in the
\disalg~ group were supplied two configurations of the \disalg~ summaries. One
summary where the first agent is the Leader while the second is the Disagreer,
and the opposite summary, where the first agent is the Disagreer and the second
is the Leader. 
Upon
conclusion, participants answered a series of explanation
satisfaction questions adapted from \cite{hoffman2018metrics}. 

\emph{Evaluation Metrics and Analyses.} 
The main evaluation metric of interest was the success rate of identifying the superior Frogger agent with each summary method. We compare this metric across all the agent selection tasks given to participants. We also compare participants’ confidence in their decision.
To compare the explanation satisfaction ratings given to the summaries, we
averaged the values of the different items normalizing in such a way that
higher values always mean that the summary is more helpful.
In all analyses we used the non-parametric Mann-Whitney $U$ test and computed
effect sizes using rank-biserial correlation. In all plots the error bars depict
the bootstrapped $95\%$ confidence intervals~\cite{efron1994introduction}. 

\begin{figure*}[t]
	\centering
	\includegraphics[width=0.85\linewidth]{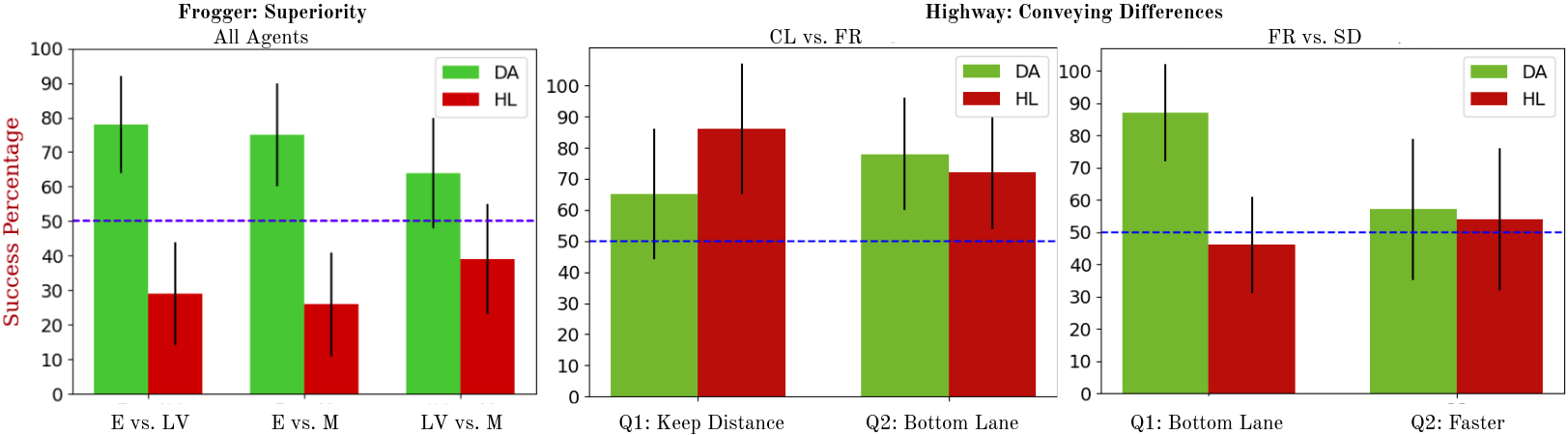}\\
	\caption{Participant Success Percentage per Summary Method. \textbf{DA:} DISAGREEMENTS; \textbf{HL:} HIGHLIGHTS.}
	\label{fig: DA HL compare}
\end{figure*}

\paragraph{Experiment 2 - Conveying Agent Differences}
This experiment's objective was to test the usefulness of \disalg~ summaries for conveying differences in general agent behavior in comparison to HIGHLIGHTS.

\emph{Hypotheses.} 
The \disalg~ algorithm is designed to portray instances of disagreement between
agents. We hypothesized that this would provide a clear and contrastive distinction between the alternative agents, thus emphasizing behavioral differences and providing a more appealing visual experience.
More specifically,we state the following hypotheses:\\ 
\textbf{H3:} Success rate of participants shown \disalg~  summaries will surpass that of participants shown HIGHLIGHTS summaries.  
\\
\textbf{H4:} participants will prefer summaries generated by the \disalg~ method.

\emph{Experimental Conditions}
A within subject setup was chosen in order to allow participants to provide a
direct comparison between the methods and state their preferences. As
participants experience both HIGHLIGHTS and \disalg~ summaries, to reduce cognitive overload, we chose to display only two agent comparisons for each
method. We chose the comparisons between the most distinctly dissimilar agents,
dropping $\langle SD,CL\rangle$ due to similarity of reward associated with
distance from neighboring vehicles. 

\emph{Participants.} 45 participants were recruited through Amazon Mechanical
Turk (13 female, mean age $= 37.51$, STD $= 10.35$), receiving similar pay and
bonus incentive as in the first experiment.
As in Exp\#1, participants were filtered out if their task completion time was below a threshold.

\emph{Procedure.} Participants followed a similar procedure as in the previous
experiment diverging solely in the domain introduction and the questions asked.
Instead of superiority between the agents, participants were queried about which
trait was more dominant in the compared agents. The ground truth was established
directly from the reward functions of the agents. Each comparison included two
such questions along with a mandatory confidence rating and textual explanation of the answers. In addition, participants were ultimately asked which method they preferred. Since this was a within-subject design, participants saw both methods, in a random order. All questions are provided in the Appendix.

\emph{Evaluation Metrics and Analyses.} 
The main evaluation metric of interest was the success rate of correctly assigning a more dominant trait to an agent. We compare this metric across the
different agent selection tasks and summary methods. Similarly to experiment 1, we compare participants’ confidence in their answers. For evaluating summary method satisfaction of participants we used the non-parametric Wilcoxon signed-rank test \cite{wilcoxon1947probability} for matched pairs.

%% file: results.tex
\section{Results}
We now describe the results of our comparison between the HIGHLIGHTS strategy summarization method and our novel \disalg~ approach. We report the main experimental results with respect to the hypotheses raised in the previous section. 

Figure \ref{fig: DA HL compare} shows the percentage of participants who were successful in answering the experiment tasks in each of the experiment conditions, for each agent pair combination.

\emph{(H1) Participants in the HIGHLIGHTS condition struggled to successfully identify the better
performing agent in the comparison task.} When all agents are of decent performance, we see the difficulty
of distinguishing between them manifest itself in a poor success rate.
Based on participants' textual explanations of the choice of agent, it seems they were concerned that agent $E$ was indecisive, e.g., ``[Agent $E$] seems very indecisive while ... [Agent $M$] seems to have a plan and is going with it.''. 
We hypothesize that these responses are a consequence of a single trajectory in agent $E$'s summary where the frog is seen leaping between logs in a seemingly indecisive manner. These results emphasize the limitations of independent comparisons.



\emph{(H2) Participants in the \disalg~ condition were more successful in the agent comparison
task}. Participants in the \disalg~ group 
showed vast improvement
in the ability to identify the better performing Frogger agent  (see Figure \ref{fig: DA HL compare}). The differences in success rate between conditions  were statistically significant and  
substantial for all agent comparisons ($E$ vs. $LV$: $p = 1.6^{-5}$; $E$ vs. $M$: $p = 1.7^{-5}$; $LV$ vs. $M$: $p = 0.018$). Textual explanations provide insights regarding how the
contrastive nature of the \disalg~ summaries helped participants decide which agent to choose, e.g. `` I preferred the path that ... [Agent $E$] was taking''; ``I felt that ... [Agent $E$] was making slightly stronger moves, and pushing ahead further''.

\emph{(H3) A significant difference was found between success rates of participants in the conveying differences task}. Participants achieved a significantly higher success rate with the \disalg~ method in the $FR$ vs. $SD$ comparison ($p = 0.014$). Albeit, this was mostly a result of the low performance of HIGHLIGHTS in $Q1$ due to the $SD$ summary containing, coincidentally, only trajectories of the agent at the bottom lane. 
The inferior performance of \disalg~ in $Q1$ of $CL$ vs. $FR$  ($p = 0.187$) can be explained by summary trajectories where no vehicles were present in $CL$'s lane allowing it to drive faster than $FR$ and appear less considerate of keeping distance.
While not necessarily outperforming it, the \disalg~ is at least equivalently useful as HIGHLIGHTS, which was shown to be better than random~\cite{Tobias}.


\emph{(H4) No clear participant preference towards one summary method was observed.} Most participants answered that both methods were equally beneficial. However,
more participants found \disalg~ more \emph{helpful} and containing less irrelevant information than HIGHLIGHTS, while finding the latter more \emph{pleasing}.



\emph{Confidence and satisfaction}
In both experiments no statistically significant differences were found between the confidence or satisfaction of participants in different conditions (See Appendix).



%% file: conclusion.tex
\section{Discussion and Future Work} \label{sec:conclusion} 
With the maturing of AI, circumstances which require people to choose between alternative market-available solutions, are likely to arise. The necessity of distinguishing between alternative agents becomes ever more clear. Moreover, distinguishing between policies is key for developers when analyzing different algorithms and  configurations. 

This paper presented a new approach for comparing RL agents by generating policy disagreement summaries. Experimental results show summaries help convey agent behaviour differences and improve users' ability to identify superior agents, when one exists.




As for future work, we note the following possible directions: \textit{i)} expanding \disalg~ to enable comparison of more than two agents; \textit{ii)} testing additional state and trajectory importance methods; \textit{iii)} further enhancing the diversity between trajectories in the summary, and  \textit{iv)} formulating and defining disagreement ``types'' for generating further user-specific summaries.

%% file: sections/appendix.tex
\newpage
\appendix
\appendixpage

\section{HIGHLIGHTS for High Varying Performance Agents}
The HIGHLIGHTS algorithm has been shown to increase users' ability to select the better performing
agent in the Pacman domain \cite{amir18highlights} in the Pacman domain. However, it was also observed that when agent performance did not differ by much, participants had difficulties identifying the superior agents. Before using HIGHLIGHTS as a baseline, we wanted to verify that it indeed supports users' choices when the agents' capabilities differ substantially in the Frogger domain. Therefore, we tested the following additional hypothesis:
\textbf{H5:} Participants shown summaries generated by HIGHLIGHTS for agents with a high degree of variation in their performance will be able to identify the better performing agent.

This would provide a sanity check that there was no inherent problem specifically with the
Frogger domain, where HIGHLIGHTS has not been evaluated before.

To this end, we trained two additional agents with comparably poor performance to test against the $E$ agent.
\emph{Additional Agents}: 
\begin{itemize}
    \item \textbf{Novice (N):} 200 training episodes. Default rewards. Average game score: -200.
    \item \textbf{FearWater (FW):} 2000 training episodes. Significantly higher negative reward for dying in the river. Average game score: -1082.
\end{itemize}

An additional experiment was conducted with 74 participants recruited through Amazon Mechanical Turk (33 female, mean
age $= 39.89$, STD $= 11.5$). The experiment procedure was identical to the one mentioned in the paper for the group witnessing the HIGHLIGHTS summaries, except for the change in agents shown.

\begin{figure}[ht]
	\centering
	\vspace{-0.4cm}
	\includegraphics[width=0.98\columnwidth]{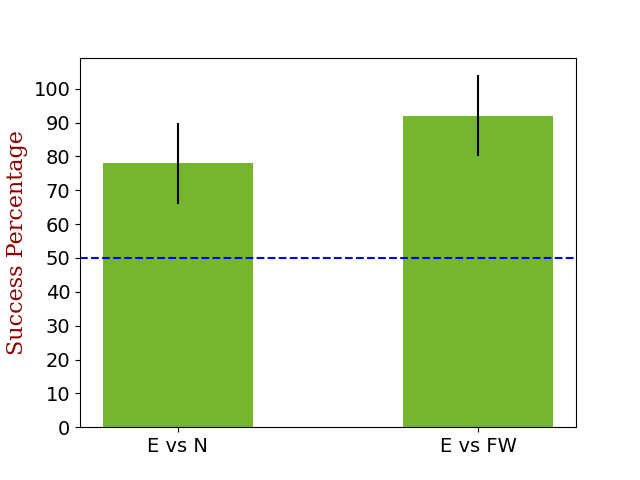}\\
	\vspace{-0.7cm}
	\caption{Success percentage in agent superiority identification task for HIGHLIGHTS summaries of agents with high-varying performance}
	\label{fig: HL easy vote}
	\vspace{-0.2cm}
\end{figure}

\textbf{Results:} \emph{(H5) Participants shown HIGHLIGHTS summaries of high-varying performance agents were able to
correctly select the most skilled agent.} Participants’ selection for this superiority identification task are shown in Figure \ref{fig: HL easy vote}. As predicted, for agents whose
performance differs greatly, the summaries produced by the HIGHLIGHTS algorithm supply sufficient information for determining which agent outperforms the other. 

\section{Summary Sensitivity to Trajectory Horizon}
\begin{figure}[ht]
	\centering
    \includegraphics[width=0.98\columnwidth]{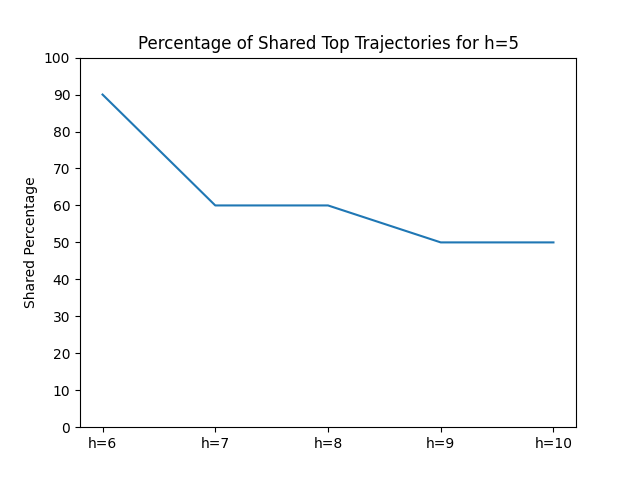}\\
	\caption{\disalg~ algorithm sensitivity to horizon parameter $h$}
	\label{fig: h_sensitivity}
\end{figure}

When choosing the trajectory horizon value ($h$), several considerations were kept in mind.
Intuitively, the further an agent is from the disagreement state, the less direct influence that
state has over the agent's current predicament; on the other hand, too low of a value may not
showcase enough of the divergence. In addition, displaying long trajectories in the user study may
lead to loss of participants' interest, while short trajectories can cause the disagreement to be
easily missed. To test the sensitivity of our algorithm to variations in $h$ we observed the top
trajectories obtained for varying values, in the Frogger domain, and noted the shared trajectories between them.
Figure \ref{fig: h_sensitivity} provides evidence that although the sensitivity exists, a
significant proportion of the trajectories stays unhindered even when doubling the value. We note
here that additional constraints that can explain the reduction in shared trajectories include
termination of the simulation and $overlapLim$.

\section{Summary Explanation Satisfaction Questions}
In each experiment participants were asked to rate the following explanation satisfaction statements on a 7-point Likert scale (0 - Strongly disagree to 6 - Strongly agree) for
the summary method they have just witnessed:
\begin{itemize}
	\item From watching the [\emph{method}] videos of the AI agents, I understood which agent is better. 
	\item The [\emph{method}] videos showing the AI agents play contain sufficient detail for deciding which agent is better. 
	\item The [\emph{method}] videos showing the AI agents play contain irrelevant details.
	\item The [\emph{method}] videos showing the AI agents play were useful for the task. 
	\item The specific scenarios shown in the [\emph{method}] videos were useful for the task. 
\end{itemize}
The questionnaire also included an attention check for detecting negligent participants.
Participants who did not answer this question correctly were not included in the analysis.

Explanation satisfaction of participants shown \disalg~ summaries was similar to that of participants shown HIGHLIGHTS summaries. Participants’ distributions of scores for the explanation satisfaction were not statistically significant $(p_{exp\#1} = 0.17, p_{exp\#2} = 0.61)$.

\begin{figure}[ht]
	\centering
    \includegraphics[width=0.98\columnwidth]{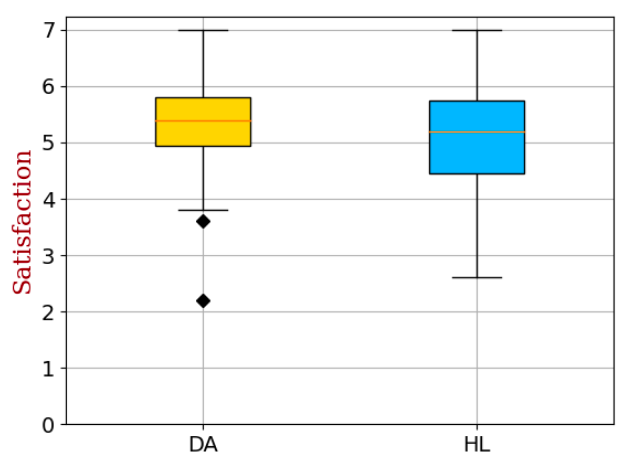}\\
	\caption{\textbf{Frogger:} Participants' Overall Summary Method Satisfaction}
	\label{fig: pref}
\end{figure}

\begin{figure}[ht]
	\centering
    \includegraphics[width=0.98\columnwidth]{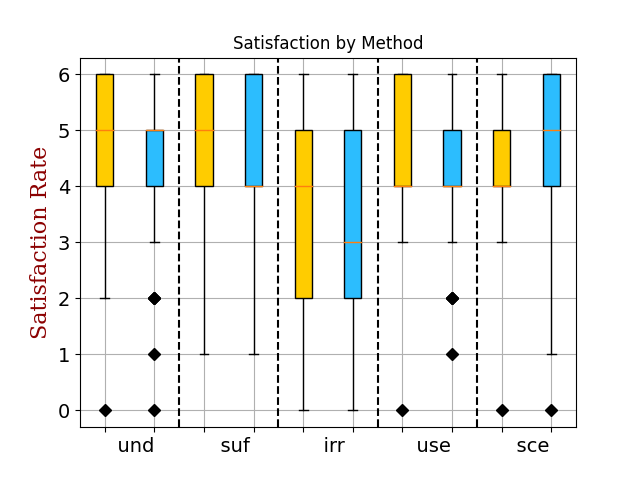}\\
	\caption{\textbf{Highway:} Participants' Summary Method Satisfaction by Question.}
	\label{fig: pref}
\end{figure}

\section{Summary Method Preference Questions}
After completing Exp\#2, participants were asked to state their preferred summary method (HIGHLIGHTS, \disalg~ or ``Both equally'') for the following attributes:
\begin{itemize}
	\item Which video method was more helpful to you in your task?
	\item Which video method was more pleasing to watch?
	\item Which video method contained more irrelevant information?
\end{itemize}
The results of this comparison are displayed in Figure \ref{fig: pref}.

\begin{figure}[ht]
	\centering
    \includegraphics[width=0.98\columnwidth]{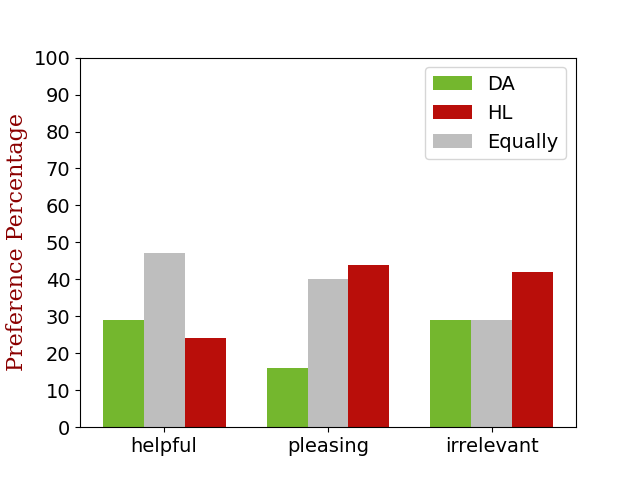}\\
	\caption{Participants' Summary Method Preference}
	\label{fig: pref}
\end{figure}

\section{Trajectory Importance Methods}
We report here additional trajectory importance methods tested in our work which
were abandoned in favor of the last-sate method which provided in our opinion
the best results. Let us denote the value of a trajectory as
$V({t}_{h}^{\pi}(s))$. 
\begin{itemize}
    \item \textbf{Sum:} 
    $V({t}_{h}^{\pi}(s)) = \sum_i V(s_{+i})$
    \item \textbf{Average:} 
    $V({t}_{h}^{\pi}(s)) = \frac{1}{h} \cdot \sum_i V(s_{+i})$
    \item \textbf{Max-Min:} 
    $V({t}_{h}^{\pi}(s)) = \max_i V(s_{+i}) -  \min_i V(s_{+i})$
    \item \textbf{Max-Avg:}
    $V({t}_{h}^{\pi}(s)) = \max_i V(s_{+i}) - \frac{1}{h} \cdot \sum_i
    V(s_{+i})$
    \item \textbf{Sum Delta:}
    $V({t}_{h}^{\pi}(s)) =  \sum^{h-1}_i \big (V(s_{+i}) - V(s_{+i+1}) \big) $ 
 
\end{itemize}
The above methods calculate $V(t)$ which is then used for calculating the
trajectory importance in the following manner:
$ Im({t}_{h}^{\pi_{L}}(s),{t}_{h}^{\pi_{D}}(s)) = 
|V(t_{h}^{\pi_L})(s) - V(t_{h}^{\pi_D}(s))| $

\section{Experiment Survey, Summary Videos and Code}
A PDF file with the principal sections of the experiment surveys, summary videos and the
code used for DISAGREEMENTS and HIGHLIGHTS are provided in the supplementary
files. 